\documentclass[runningheads]{llncs}
\usepackage[T1]{fontenc}
\usepackage{graphicx}
\usepackage{color}
\usepackage{amsmath}
\usepackage{makecell}
\usepackage{booktabs}
\usepackage{algorithm}
\usepackage{algorithmic}
\usepackage{subfigure}
\usepackage{adjustbox}
\usepackage{wrapfig}
\usepackage{colortbl}
\usepackage{arydshln}

\usepackage{array}   

\definecolor{newpink}{RGB}{240,151,153}
\definecolor{neworange}{RGB}{247,203,153}
\definecolor{newyellow}{RGB}{255,255,153}

\begin{document}
\title{Learning Segmented 3D Gaussians via Efficient Feature Unprojection
for Zero-shot Neural Scene Segmentation}
%
%
\author{
Bin Dou\inst{1}
\and
Tianyu Zhang\inst{1}
\and
Zhaohui Wang\inst{1}
\and
Yongjia Ma\inst{1}
\and
Zejian Yuan\inst{1}
}
%
%
\institute{
Institute of Artificial Intelligence and Robotics, Xi’an Jiaotong University\\
\email{
\{abc991227db, puzzling229, astjkl12345, maguire\}@stu.xjtu.edu.cn, yuan.ze.jian@xjtu.edu.cn
}
}
\maketitle              
\begin{abstract}

Zero-shot neural scene segmentation, which reconstructs 3D neural segmentation field without manual annotations, serves as an effective way for scene understanding. 
However, existing models, especially the efficient 3D Gaussian-based methods, struggle to produce compact segmentation results.
This issue stems primarily from their redundant learnable attributes assigned on individual Gaussians, leading to a lack of robustness against the 3D-inconsistencies in zero-shot generated raw labels. To address this problem, our work, named \textbf{Co}mpact \textbf{Seg}mented 3D \textbf{Gaussians} (\textbf{CoSegGaussians}), proposes the Feature Unprojection and Fusion module as the segmentation field, which utilizes a shallow decoder generalizable 
for all Gaussians based on high-level features. Specifically, leveraging the learned Gaussian geometric parameters, semantic-aware image-based features are introduced into the scene via our unprojection technique. The lifted features, together with spatial information, are fed into the multi-scale aggregation decoder to generate segmentation identities for all Gaussians. Furthermore, we design \textit{CoSeg Loss} to boost model robustness against 3D-inconsistent noises. Experimental results show that our model surpasses baselines on zero-shot semantic segmentation task,
improving by $\sim10$\% mIoU over the best baseline. 
Code and more results will be available
at https://David-Dou.github.io/CoSegGaussians.

\keywords{
Neural Scene Segmentation
\and
Novel View Segmentation
\and
3D Gaussian Splatting
\and
Zero-shot Segmentation.
}

\end{abstract}

\section{Introduction}

In recent years, Neural Radiance Field (NeRF)~\cite{mildenhall2021nerf} and its subsequent methods
~\cite{barron2022mip,fridovich2022plenoxels,kerbl20233d} 
have propelled the development of neural scene representation and rendering.
Based on the technique, 
many methods\cite{wang2022dm,zhi2021place} have emerged to achieve neural scene segmentation, i.e., lifting 2D posed information into 
3D neural representation
for scene segmentation.
High-quality segmentation methods
form the foundation for scene understanding and editing, which are essential to applications such as robot navigation, automatic driving and VR/AR~\cite{siddiqui2023panoptic}.


\begin{figure}[ht]
    \centering
    \begin{adjustbox}{center}
    \begin{minipage}{0.375\textwidth}
      \centering
      \includegraphics[width=\linewidth]{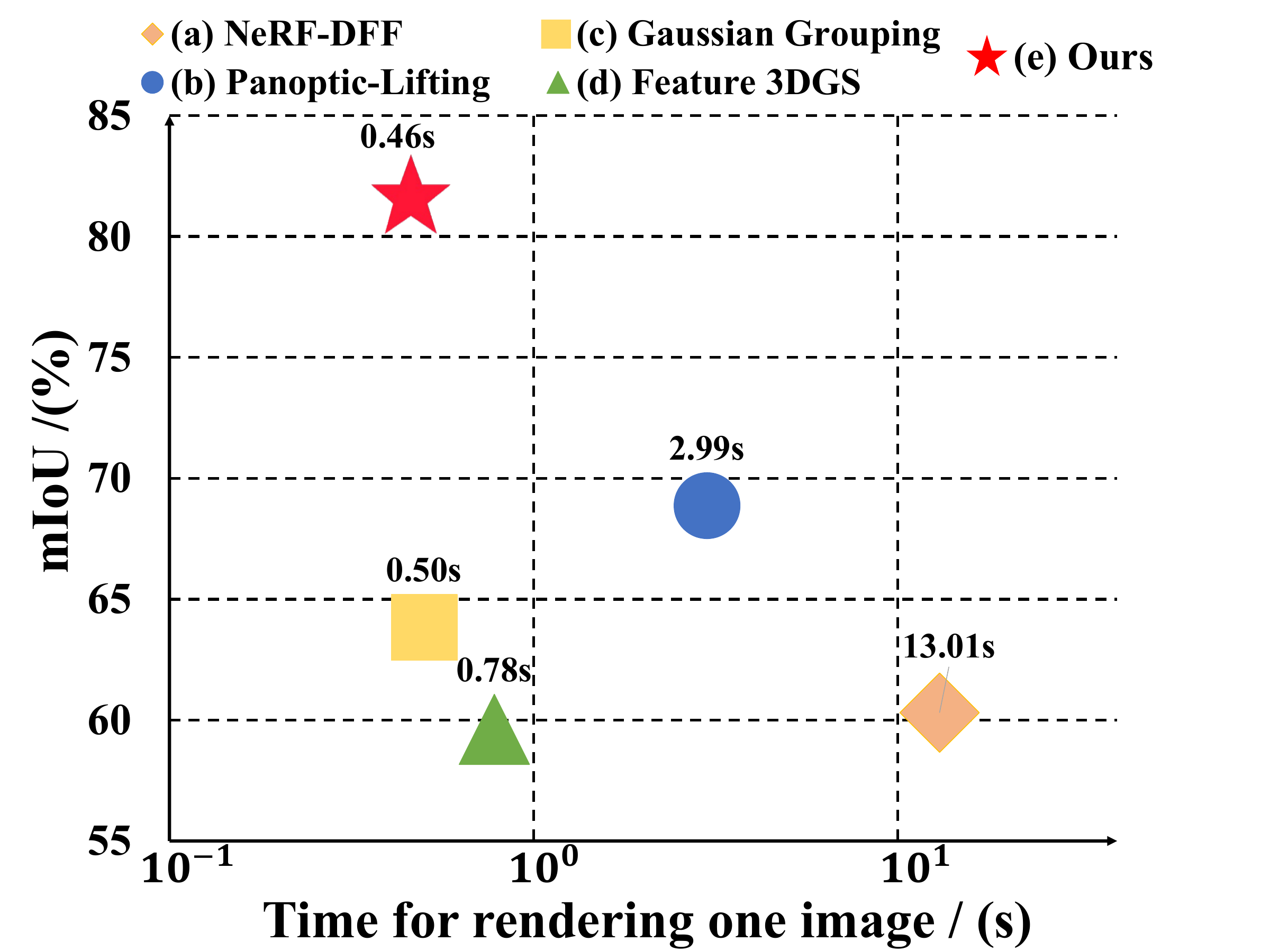} 
      \caption{
      Comparison of performance and inference speed for zero-shot scene segmentation.}
      \label{fig:startup}
    \end{minipage}%
    \hfill
    \hspace{0.3cm} 
    \begin{minipage}{0.635\textwidth}
      \centering
      \includegraphics[width=\linewidth]{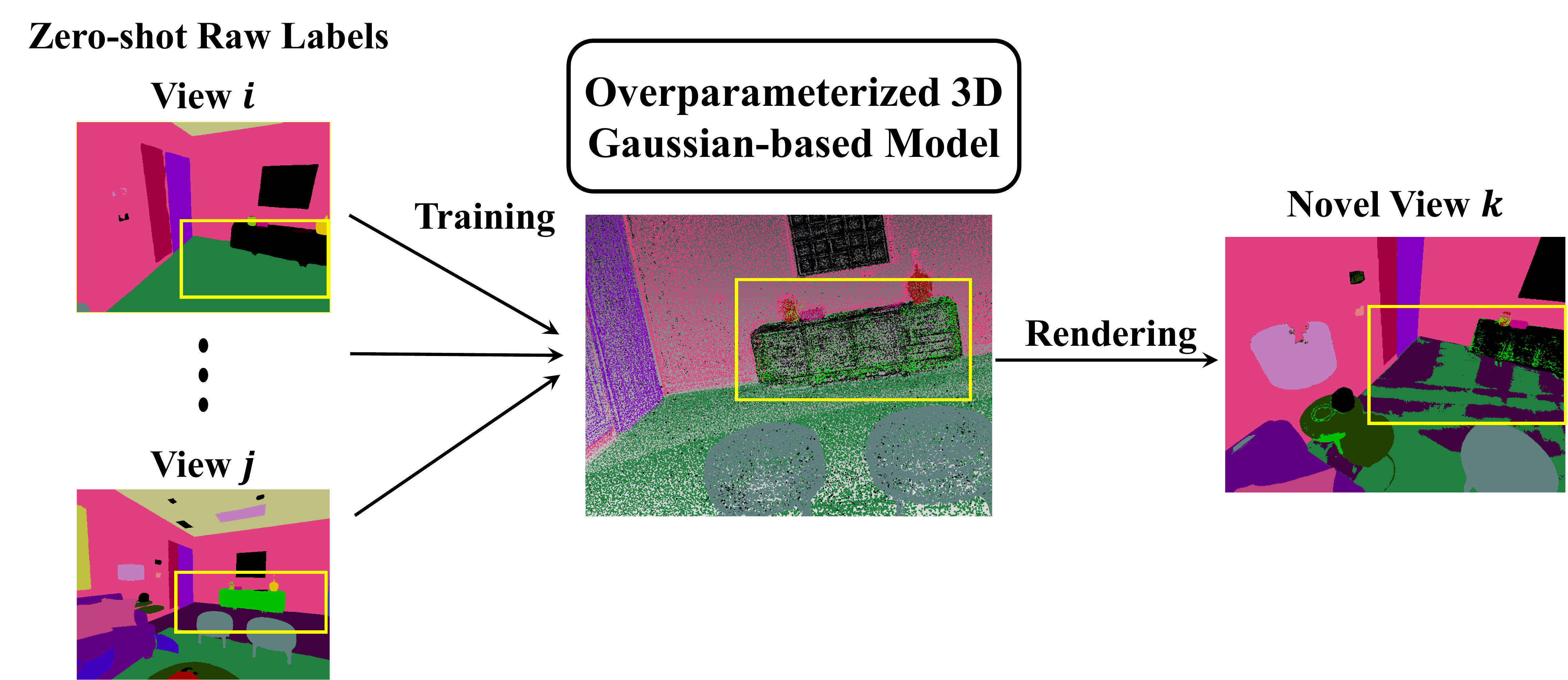} 
      \caption{
      Failure for compact 3D segmentation caused by cross-view inconsistency (\textcolor[rgb]{0.8,0.8,0.0}{selected regions}) in previous Gaussian-based method.
      }
      \label{fig:reason}
    \end{minipage}
    \end{adjustbox}
\end{figure}

Early neural scene segmentation approaches~\cite{wang2022dm,zhi2021place} have relied on annotation for the given scene such as dense-view 3D-consistent labeled masks,
which are time-consuming to acquire.
Aiming to reduce reliance on manual annotations and improve the cross-scene transferability, zero-shot neural scene segmentation becomes the research trend~\cite{siddiqui2023panoptic,ye2023gaussian}, which means achieving segmentation without any 
manual annotations for the scene and is the task of our study.

To accomplish the zero-shot segmentation,
recent works~\cite{kobayashi2022decomposing,siddiqui2023panoptic,ye2023gaussian,zhou2023feature} have applied image-based semantic-aware knowledge or labels generated from pre-trained vision models, such as DINO~\cite{caron2021emerging}, Mask2Former~\cite{cheng2022masked} and SAM~\cite{kirillov2023segment}. 
However, these methods exhibit low performance, as shown in Fig. \ref{fig:startup}.
While approaches merely using high-level features (Fig. \ref{fig:startup} (a \& d)) lack effective identity mapping mechanism for the zero-shot visual task, methods using raw labels also demonstrate underperformance,
especially those~\cite{ye2023gaussian} based on the efficient 3D Gaussian Splatting representation (Fig. \ref{fig:startup} (c)).
This issue mainly stems from the overparameterization of those models, which directly assign redundant learnable attributes on individual Gaussians, resulting in overfitting to cross-view inconsistent noises in zero-shot generated masks. As shown in Fig. \ref{fig:reason}, for the 3D object or region where different labels are provided on different viewpoints, existing Gaussian-based methods with heavily redundant learnable attributes will overfit to the inconsistency and fail to achieve compact segmentation, i.e., providing the same label inside semantic-level one 3D object~\cite{siddiqui2023panoptic}.

In this paper, we present CoSegGaussians, a Gaussian-based approach for efficient and compact zero-shot neural scene segmentation (see Fig. \ref{fig:startup} (e)).
Unlike the redundant learnable attributes on each primitive, we design the decoder generalizable for all Gaussians to represent the segmentation field. 
The decoding network takes high-level (semantic-aware and spatial) features as inputs, and fuses them to obtain segmentation identities, enabling the network to represent segmentation field with only shallow layers.
For lifting image-based semantic-aware features from 2D foundation models into scene representation, we also propose an efficient unprojection method. 
It performs as an explicit inverse rendering method 
across training viewpoints,
which avoids the time-consuming high-dimensional rendering, problem~\cite{zhou2023feature} (as the feature dimension increases, Gaussian rasterization speed drops drastically) and significantly reduces learnable parameters in previous feature distillation methods.
By addressing the cross-view inconsistency in zero-shot generated masks as label noises, our work is the first to employ pixel-wise label noise robustness loss for the zero-shot task. Through incorporation with 
3D regularization, the segment compactness loss guides segmentation field learning in the presence of label inconsistencies.

In extensive experiments, it's illustrated that our method delivers segmentation quality surpassing previous zero-shot methods, shown in the performance improvement of novel view segmentation. 
At model efficiency, our method achieves faster inference speed (especially compared to NeRF-based methods) and requires less learnable parameters than existing Gaussian-based methods.
Furthermore, we investigate our unprojection technique for building fields of features from two different pre-trained foundation models, DINO~\cite{caron2021emerging} (visual semantic-aware model) and LSeg~\cite{li2022language} (language-driven segmentation model), showing the effectiveness of our method for Gaussian feature field reconstruction.


In summary, our work makes the following contributions:

$\bullet$
We propose the Feature Unprojection and Fusion module to 
learn 3D Gaussian segmentation field, 
which addresses the model overparameterization problem
and facilitates compact zero-shot neural scene segmentation.

$\bullet$
We present the unprojection technique for introducing 2D posed features into 3D Gaussians, 
which enables efficient Gaussian feature field reconstruction.

$\bullet$
We design \textit{CoSeg Loss} to supervise the
segmentation field learning, which enhances the robustness against 3D-inconsistencies in zero-shot generated labels.

\section{Related Work}

\textbf{Neural Scene Representation. }
Neural scene representation has seen significant advancements
in recent years.
NeRF~\cite{mildenhall2021nerf} encodes a 3D scene's geometry and appearance with MLPs, combining 
volume rendering 
for
neural scene optimization and novel view synthesis. 
Subsequent works have introduced numerous improvements, 
as 
~\cite{barron2022mip,hu2023tri,yu2021pixelnerf,zhang2020nerf++}
aim to enhance rendering quality, 
while ~\cite{chen2022tensorf,chen2023mobilenerf,fridovich2022plenoxels,muller2022instant,sun2022direct}
focus on acceleration.
Recently, 3D Gaussian Splatting (3DGS)~\cite{kerbl20233d}
garners widespread attention due to its capability 
upon
high-quality and real-time rendering.



\textbf{Scene Segmentation. }
For 3D scene understanding, numerous approaches have extended the neural 
radiance field
to segmentation field. 
Semantic-NeRF~\cite{zhi2021place} and DM-NeRF~\cite{wang2022dm} achieve neural scene segmentation by 
concurrent optimization of the radiance and segmentation field, relying on dense-view 3D-consistent annotated labels. 
Benefiting from the 
pre-trained 2D foundation models,
some methods have been developed for zero-shot segmentation.
Panoptic-Lifting~\cite{siddiqui2023panoptic} 
achieves scene segmentation by utilizing generated labels, while PanopticNeRF~\cite{fu2022panoptic} additionally employ 
coarse 3D labels. 
Gaussian Grouping~\cite{ye2023gaussian} extends 3DGS by assigning segmented identities on each Gaussian, 
and employ SAM~\cite{kirillov2023segment}-generated and further associated masks for supervision.
Feature 3DGS~\cite{zhou2023feature} introduces knowledge from 2D foundation models into 3D Gaussian representation via feature distillation and can further achieve novel view segmentation.

Considering the point-like characteristic of 3D Gaussians, it's hypothesized that modules in point cloud segmentation can be extended 
for Gaussian-based methods.
PointNet++~\cite{qi2017pointnet++}, 
SPG~\cite{landrieu2018large} 
can be used for scene point cloud segmentation but are computationally expensive. 
RandLA-Net~\cite{hu2020randla} employs 
rapid random sampling combined with Local Feature Aggregation 
to improve model efficiency, particularly for large-scale point clouds.
Trading-off performance and efficiency, 
we try to utilize spatial information from RandLA-Net for 3D Gaussians.

\textbf{Label Noise Robustness Loss. } 
By treating the cross-view inconsistencies in generated masks as label noises, 
pixel-wise label noise robustness objective is designed. 
Previous loss functions, such as
MAE~\cite{ghosh2017robust}, 
RCE~\cite{ma2020normalized}, 
GCE~\cite{zhang2018generalized}
and GJS~\cite{englesson2021generalized} are proposed 
to replace the cross-entropy loss
for label-noise-conditioned image classification task.
In CoSegGaussians, we extend JS loss to pixel-wise situation 
against cross-view inconsistent noises in the zero-shot raw labels.

\section{Method}

\begin{figure*}[!ht]
    \centering
    \begin{adjustbox}{center}
        \includegraphics[width=1.02\linewidth]{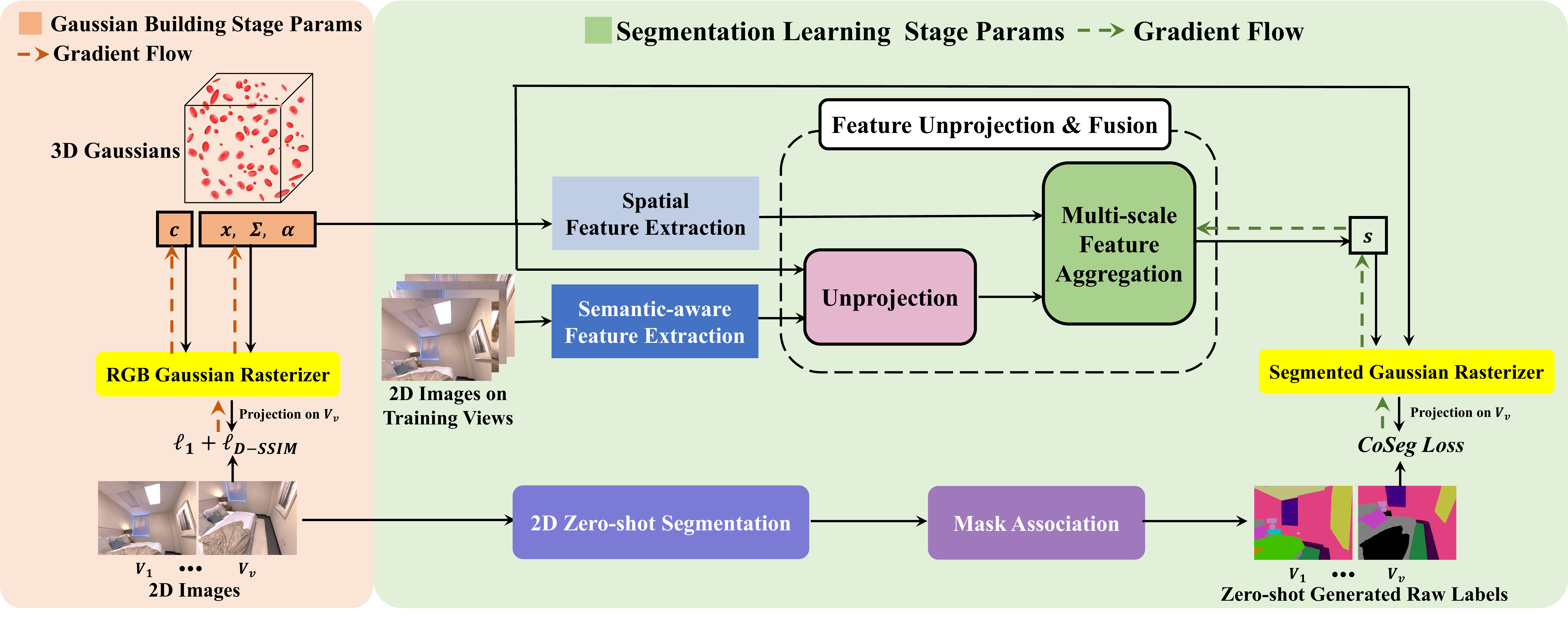}
    \end{adjustbox}
    \caption{
    Overview of our method.
    Training process takes a two-stage strategy.
    In Gaussian Building Stage, geometric ($\boldsymbol{x, \Sigma, \alpha}$) and appearance ($\boldsymbol{c}$) attributes are optimized following 3DGS~\cite{kerbl20233d}.
    Next in Segmentation Learning Stage, segmentation identities $\boldsymbol{s}$ are learned with the Feature Unprojection \& Fusion module. It takes the extracted 3D spatial information as well as the image-based semantic-aware features as input, as the latter is introduced into 3D Gaussians efficiently via our unprojection technique. 
    After fusing the above features with the multi-scale aggregation network, $\boldsymbol{s}$ for all Gaussians are generated, which are supervised by comparing the rendered segmentation map with the zero-shot segmented and further associated raw label, guided by our \textit{CoSeg} Loss.
    For inference, with Gaussian attributes $\boldsymbol{x, \Sigma, \alpha}$ and $\boldsymbol{s}$ learned, segmentation maps are rendered on novel views.
    }
    \label{fig:model}
\end{figure*}

Given only posed RGB images of a 3D scene, our method aims to build an expressive representation to capture the scene's geometry, appearance
together with the high-quality segmentation identities.
The proposed model, CoSegGaussians, 
enables compact neural scene segmentation, generating segmentation from novel views in a 3D-consistent fashion.
Fig. \ref{fig:model} provides an overview of our method.
We first introduce 3D Gaussian representation in Sec. \ref{preliminary}.
Subsequently, Sec. \ref{fea_fusion} presents Feature Unprojection \& Fusion module for Gaussian segmentation field learning. Sec. \ref{train} elaborates the training strategy.

\subsection{3D Gaussian Scene Representation}
\label{preliminary}

We adopt the efficient 3D Gaussian Splatting~\cite{kerbl20233d} 
to represent the scene.
Each Gaussian primitive is parameterized by its centroid position 
$\boldsymbol{x}$,
3D covariance $\boldsymbol{\Sigma}$, 
opacity 
$\alpha$
and view-dependent color $\boldsymbol{c}$. 
$\boldsymbol{x}, \boldsymbol{\Sigma}, \alpha$ are regarded as the geometric attributes 
while $\boldsymbol{c}$ denotes the appearance.
To optimize these learnable parameters, 
3DGS projects them onto the 2D imaging plane to render the RGB image 
by $\alpha$-blending, 
a differentiable point-based rasterization technique.
For each pixel $\boldsymbol{p}$, the 
$\alpha$-blending is formulated as:
\begin{equation}
    \label{color}
    \boldsymbol{C}(\boldsymbol{p}) = \sum_{i \in \mathcal{N}} \boldsymbol{c}_i \alpha_i'(\boldsymbol{p}) \prod_{j=1}^{i-1}(1 - \alpha_j'(\boldsymbol{p}))
\end{equation}
where $\mathcal{N}$ denotes the ordered Gaussians overlapping the pixel, $\boldsymbol{c}_i$ represents the $i$-th Gaussian's color and $\alpha_i'(\boldsymbol{p})$ 
expresses 
the projected opacity on $\boldsymbol{p}$,
obtained by multiplying the projected 2D covariance with the learned 
$\alpha_i$.
By defining the Gaussian projection function for 2D opacity as $\alpha_{2D}$, $\alpha_i'(\boldsymbol{p})$ can be expressed as:
\begin{equation}
    \label{alpha}
    \alpha_i'(\boldsymbol{p})= \alpha_{2D}(x_i, \boldsymbol{\Sigma}_i, \alpha_i, \boldsymbol{p})
\end{equation}

To achieve scene segmentation, 
we introduce an additional view-independent attribute, Segmentation Identity $s$, to each Gaussian.
By extending the RGB rasterizer 
to the segmentation field, identity $\boldsymbol{S}$ on pixel $\boldsymbol{p}$ can be obtained by:
\begin{equation}
    \label{sem}
    \boldsymbol{S}(\boldsymbol{p}) = \sum_{i \in \mathcal{N}} \boldsymbol{s}_i \alpha_i'(\boldsymbol{p}) \prod_{j=1}^{i-1}(1 - \alpha_j'(\boldsymbol{p}) )
\end{equation}


\subsection{Feature Unprojection \& Fusion}
\label{fea_fusion}

In this section, we describe Feature Unprojection \& Fusion module for Gaussian segmentation attribute learning,
following the order as \textbf{image-based feature unprojection} and \textbf{multi-scale feature fusion}.

\subsubsection{Image-based Feature Unprojection}
\label{unprojection}


To introduce the 2D posed feature maps extracted from RGBs into the scene representation efficiently, we design the unprojection, an explicit inverse rendering method, i.e., without learning-based 3D distillation, which avoids the time-intensive high-dimensional feature rendering and eliminates the need for redundant learnable feature attributes.

Our method is developed from the core perspective of distillation-based feature field reconstruction, as the 2D feature map extracted by the foundation model can be treated as reconstructed 3D feature field's rendering, obtained by:
\begin{equation}
    \label{color}
    \boldsymbol{\mathit{F}}_{{d}_{n}}(\boldsymbol{p}) 
    = \sum_{i \in \mathcal{N}} \boldsymbol{\mathit{f}}_{{d}_{n},{i}}\alpha_i'(\boldsymbol{p}) \prod_{j=1}^{i-1}(1 - \alpha_j'(\boldsymbol{p}))
\end{equation}
where $\boldsymbol{\mathit{F}}_{{d}_{n}}(\boldsymbol{p})$ is the $n$-th scale's feature (extracted feature from the $n$-th foundation model block) on pixel $\boldsymbol{p}$, $\boldsymbol{\mathit{f}}_{{d}_{n},{i}}$ is the corresponding feature on the $i$-th Gaussian.
Through employing learnable $\boldsymbol{\mathit{f}}_{{d}_{n},{i}}$ and 
comparing the rendering with the extracted feature map (as target), 
previous methods gather features across training viewpoints to reconstruct the 3D feature field.


It can be noticed in Eq. \ref{alpha} that with the 3D Gaussians' geometric attributes ($\boldsymbol{x}$, $\boldsymbol{\Sigma}$, $\alpha$) learned, $\alpha_i'(\boldsymbol{p})$ will also be fixed for each $\boldsymbol{p}$,
leading to the fixed transmittance $T_i(\boldsymbol{p})=\prod_{j=1}^{i-1}(1 - \alpha_j'(\boldsymbol{p}))$.
Hence obtaining each Gaussian's feature from 2D maps can be turned into solving the linear equation system. 
Through deducing the solution approximately, and also inspired by ~\cite{chen2023gaussianeditor},
the $n$-th scale's feature on the $i$-th Gaussian $\boldsymbol{\mathit{f}}_{{d}_{n},{i}}
$ can be expressed as: 
\begin{equation}
    \boldsymbol{\mathit{f}}_{{d}_{n},{i}} = \frac { \sum_{v \in V} \sum_{\boldsymbol{p}_{v}} \alpha_i'(\boldsymbol{p}_{v}) * 
    T_i(\boldsymbol{p}_{v})
    * \boldsymbol{\mathit{F}}_{{d}_{n}}(\boldsymbol{p}_{v}) }
    {\sum_{v \in V}|\boldsymbol{p}_{v}|+\epsilon}
    \label{eq:unprojection}
\end{equation}
where $V$ represents all training viewpoints,
$\boldsymbol{p}_{v}$ represents the pixel on view $v$ which is affected by the $i$-th Gaussian,
$\alpha_i'(\boldsymbol{p}_{v})$ and $T_i(\boldsymbol{p}_{v})=\prod_{j=1}^{i-1} (1 - \alpha_j'(\boldsymbol{p}_{v}))$ denote opacity and transmittance on $\boldsymbol{p}_{v}$, $\boldsymbol{\mathit{F}}_{{d}_{n}}(\boldsymbol{p}_{v})$ is the $n$-th scale's feature vector on $\boldsymbol{p}_{v}$ and $\epsilon$ is the small constant for avoiding exceptions.

\begin{wrapfigure}{r}{0.55\textwidth} 
  \centering
  \includegraphics[width=\linewidth]{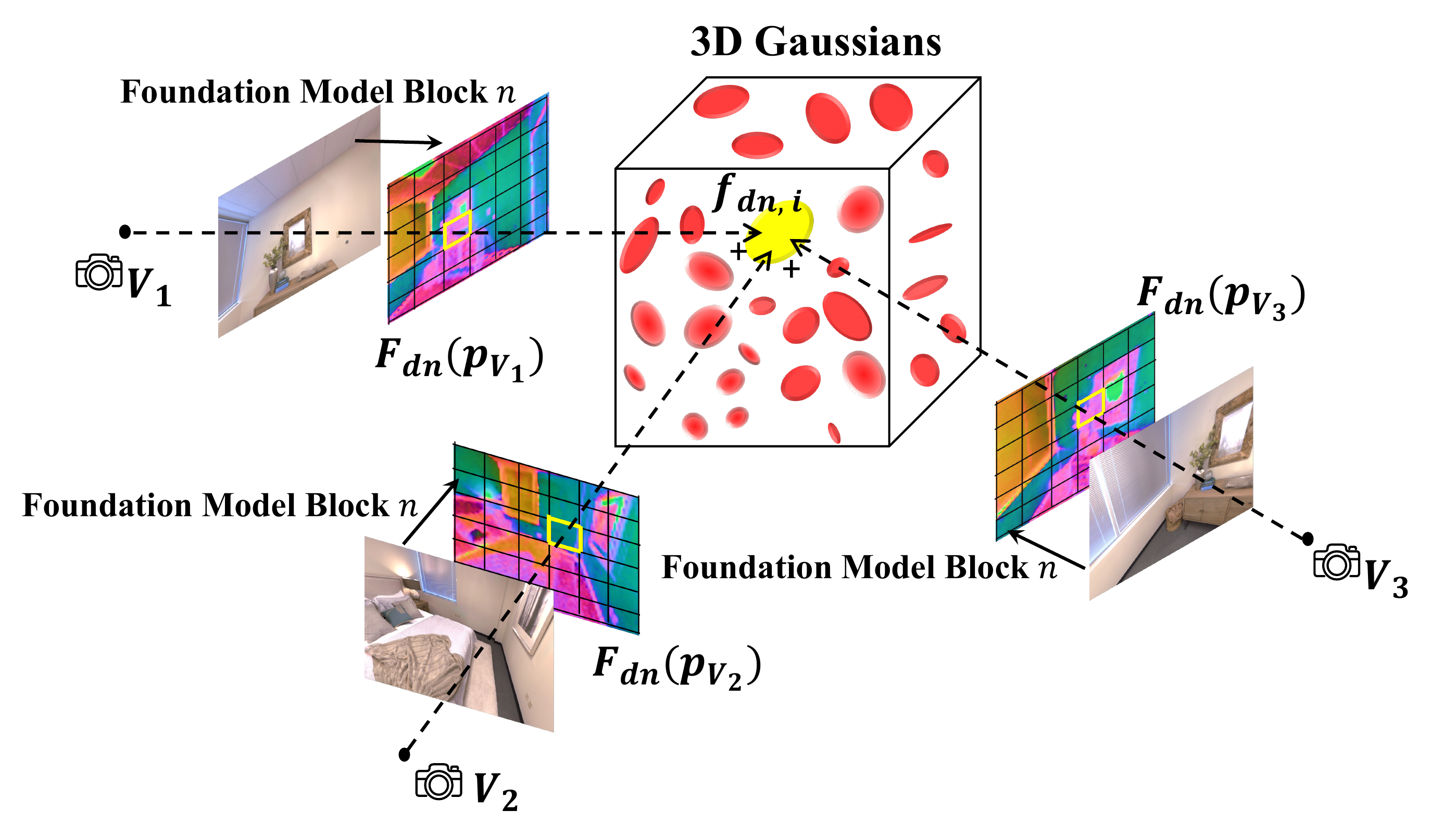} 
  \caption{Feature unprojection.
    3D Gaussian $i$'s feature can be acquired by gathering 2D features on pixels from training viewpoints if the pixel is affected by $i$.}
  \label{fig:unprojection}
\end{wrapfigure}
Based on Eq. \ref{eq:unprojection}, we design the unprojection method to lift 2D posed features into each Gaussian, which serves as an explicit
inverse rendering approach based on
weighted summation performing on all training views' pixels if the pixel is affected by the $i$-th Gaussian and further averaged over all the affected pixels,
as illustrated in Fig \ref{fig:unprojection}.


In practice,
cover \texttt{radii} in 3DGS implementation ~\cite{kerbl20233d} is applied to judge whether the pixel is affected by certain Gaussian,
and a feature buffer along with a counter
is maintained for all Gaussians.
After updating 
over all training viewpoints,
we also set attendance rate threshold to prune Gaussians adaptively according to the counter, which will improve efficiency
for further feature processing. 
Therefore, image-based feature field reconstruction on 3D Gaussians can be expressed as Algorithm \ref{alg:unprojection}. 
It can be noticed that our unprojection-based feature field reconstruction serves as an effective method for introducing image-based features extracted by foundation models, 
such as DINO~\cite{caron2021emerging} (here we select it as the visual semantic-aware feature extractor for further segmentation) and LSeg~\cite{li2022language} (language-driven model), demonstrated in Sec. \ref{sec:unprojection}.

\begin{algorithm}[H]
    \caption{Unprojection-based Feature Field Reconstruction}
    \label{alg:unprojection}
    \textbf{Input}: 2D images $\boldsymbol{C}$ on training viewpoints $V$, Corresponding feature maps $\boldsymbol{\mathit{F}}_{{d}_{n}}$
    \\
    \textbf{Parameter}: 
    Attendance Rate Threshold $\mathcal{T}$, Feature Map Size $(H, W)$, Stabilizer $\epsilon$
    \\
    \textbf{Output}: 3D Gaussian feature field $\boldsymbol{x}, \boldsymbol{\Sigma}, \alpha, \boldsymbol{\mathit{f}}_{{d}_{n}}$
    \begin{algorithmic}[1] 
    \STATE Learn $\boldsymbol{x}_i, \boldsymbol{\Sigma}_i, \alpha_i$ using  $\boldsymbol{C}$,  $i=1\cdot\cdot\cdot\mathcal{N}$ where $\mathcal{N}$ is the amount of learned Gaussians.
    \STATE Initialize $Counter=[0]_{\mathcal{N}}$, $Feature\_Buffer=[\mathbf{0}]_{\mathcal{N}}$
        \FOR{$v=1$ to $V$}
        \FOR{$\boldsymbol{p}_v=1$ to $H*W$}
        \IF {Gaussian $i$'s \texttt{radii} on $v$ covers  $\boldsymbol{p}_v$}
        \STATE $\alpha_i'(\boldsymbol{p}_{v})=\alpha_{2D}(x_i, \boldsymbol{\Sigma}_i, \alpha_i, \boldsymbol{p}_{v}), \quad T_i(\boldsymbol{p}_{v})=\prod_{j=1}^{i-1} (1 - \alpha_j'(\boldsymbol{p}_{v}))$
        \STATE $Counter$[$i$] += 1
        \STATE $Feature\_Buffer$[$i$] += $\alpha_i'(\boldsymbol{p}_{v})*T_i(\boldsymbol{p}_{v})*\boldsymbol{\mathit{F}}_{{d}_{n}}(\boldsymbol{p}_v)$
        \ENDIF
        \ENDFOR
    \ENDFOR
    \STATE Prune Gaussian $i$ where $Counter$[$i$] \textless $\mathcal{T}*V*H*W$
    \STATE $\boldsymbol{\mathit{f}}_{{d}_{n}}=Feature\_Buffer/(Counter+\epsilon)$
    \STATE \textbf{return} $\boldsymbol{x}, \boldsymbol{\Sigma}, \alpha, \boldsymbol{\mathit{f}}_{{d}_{n}}$
    \end{algorithmic}
\end{algorithm}


\subsubsection{Multi-scale Feature Fusion}
\label{sem_gen}

\begin{figure*}[!h]
    \centering
    \begin{adjustbox}{center}
    \includegraphics[width=1.0\linewidth]{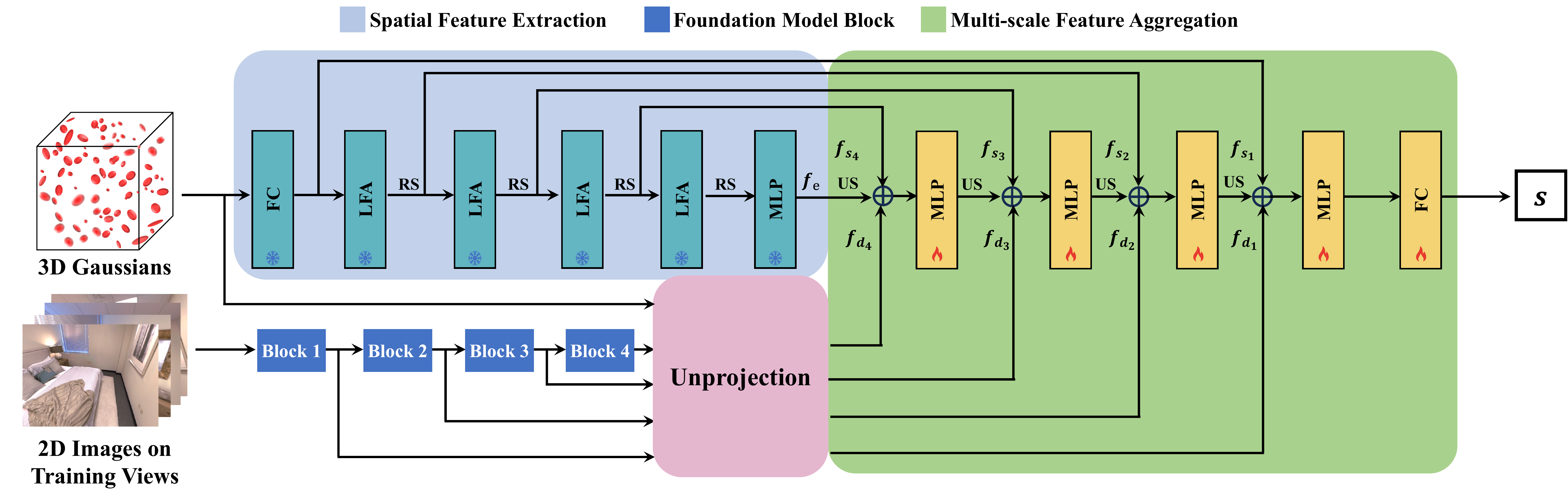}
    \end{adjustbox}
    \caption{
    Multi-scale feature fusion for segmentation identity generation.
    (\textbf{LFA}: Local Feature Aggregation, 
    \textbf{RS} \& \textbf{US}: Random \& Up Sampling, \textbf{FC}: Fully Connected Layer)
    }
    \label{fig:fusion}
\end{figure*}

Fusing unprojected image-based semantic-aware features with spatial geometric features through multi-scale aggregation generates segmentation identity $s$ for each Gaussian, as illustrated in Fig. \ref{fig:fusion}.

The spatial information of Gaussians serves as the geometric prior for scene segmentation.
To extract spatial features for point-like 3D Gaussians,
we utilize the encoders of RandLA-Net~\cite{hu2020randla}, an efficient method for segmenting large-scale point clouds, as the pre-trained extractor. 
Concretely, 
Local Feature Aggregation ($\boldsymbol{LFA}$) blocks (consisting of Local Spatial Encoding and Attentive Pooling) and Random Sampling ($\boldsymbol{RS}$) from pre-trained RandLA-Net
are employed 
on Gaussian centroids $\boldsymbol{x}$,
and the multi-scale spatial features $\boldsymbol{\mathit{f}}_{{s}_{n}}$ from local to global ($n=1$ to $4$) for sampled Gaussian points are:
\begin{equation}
    \boldsymbol{\mathit{f}}_{{s}_{n}} = \boldsymbol{RS}(\boldsymbol{LFA}_{n}(\cdot \cdot \cdot \boldsymbol{RS}(\boldsymbol{LFA}_{1}(\boldsymbol{x}))))
\end{equation}

As further rendering-based optimization only depends on 
each view's visible Gaussians 
and RandLA-Net releases the strict requirement for inputting point cloud of complete objects
due to its attention module, 
only visible Gaussians from each viewpoint
(selected 
identical to
3DGS)
instead of the whole scene are 
fed
to the spatial feature extractor 
for model efficiency. 

For the following segmentation identity generation, 
a decoder generalizable for all Gaussians is applied, 
which integrates the multi-scale unprojected semantic-aware and extracted spatial features using the global-to-local architecture. 
Specifically,
four MLP decoding layers are used to progressively propagate features from subsamplings to the originally built Gaussians.
For each layer, 
the Gaussian feature set from previous layer is upsampled through a distance-based KNN interpolation.
Then the upsampled features are fused with the corresponding scale's unprojected features $\boldsymbol{\mathit{f}}_{{d}_{n}}$ and extracted spatial information $\boldsymbol{\mathit{f}}_{{s}_{n}}$ via skip connections, operated by feature channel concatenation,
followed by
a shared MLP applied to the concatenated features.
After the global($n=4$)-to-local($n=1$) aggregation, Gaussian segmentation identities are obtained using the shared FC layers, as the
logits $s$ 
can be expressed as:
\begin{equation}
    \label{decode}
    \boldsymbol{s} = FC( \boldsymbol{\mathit{f}}_{{d}_{1}} \oplus \boldsymbol{\mathit{f}}_{{s}_{1}} \oplus \boldsymbol{US} (MLP \cdot \cdot \cdot \oplus \boldsymbol{US}(MLP(\boldsymbol{\mathit{f}}_{{d}_{4}} \oplus \boldsymbol{\mathit{f}}_{{s}_{4}} \oplus \boldsymbol{US}(\boldsymbol{\mathit{f}}_{e})
    )))
\end{equation}
where $\oplus$ is the feature concatenation, $\boldsymbol{US}$ denotes Up Sampling and $\boldsymbol{\mathit{f}}_{e}$ represents features from the final output of the spatial feature encoder.

\subsection{Training Strategy}
\label{train}

In our method, segmentation field optimization requires the learned geometric information of 3D Gaussians, as 2D feature unprojection is based on the geometric attributes
and spatial feature extraction is conducted on Gaussian centroid positions. 
Therefore, our training employs a two-stage strategy: \textbf{Gaussian Building Stage} and \textbf{Segmentation Learning Stage}.

Gaussian Building Stage focuses on obtaining geometric attributes, i.e. Gaussians' centroids $\boldsymbol{x}$, covariance $\boldsymbol{\Sigma}$, opacities $\alpha$, under the supervision of posed RGB images.
It is similar to 3DGS training, as geometric parameters along with colors $\boldsymbol{c}$ are supervised with $\lambda_{1}\ell_1+(1-\lambda_{1})\ell_{D-SSIM}$ loss. 

In Segmentation Learning Stage, the shallow MLP decoder in feature aggregation (Eq. \ref{decode}) contains the only parameters to be updated,
supervised by zero-shot segmented and further associated (to enhance 3D-consistency~\cite{ye2023gaussian}) raw labels.
And we design a loss called \textit{CoSeg Loss} to guide it for compact segmentation, which contains pixel-wise label noise robustness loss and 3D regularization.

\textbf{Pixel-wise Loss.}
For the zero-shot segmentation, considering the cross-view inconsistency in the generated raw labels as 
annotation noises, we treat the 
optimization of segmentation field as the learning with noisy labels.
Therefore label noise robustness loss is designed for pixel-wise supervision.
Instead of conventional cross-entropy, we utilize the JS loss (extended from~\cite{englesson2021generalized}) to supervise all pixels of the segmentation rendering by comparing with the zero-shot generated mask,
as the pixel-wise loss on one rendered map can be formulated as:

\begin{equation}
    \ell_{pix} = \sum_{\boldsymbol{p}} \frac { \pi_{rl} D_{\mathrm{KL}} (\kappa_{rl}(\boldsymbol{p}) \| \boldsymbol{m}) + \pi_{p} D_{\mathrm{KL}} (\boldsymbol{\boldsymbol{S}(\boldsymbol{p})} \| \boldsymbol{m})} {Z}
\end{equation}
where $D_{\mathrm{KL}}$ denotes KL divergence, $\pi_{rl}$ and $\pi_{p}$ are manually set weights for the generated \textit{raw label} and prediction respectively, $\kappa_{rl}(\boldsymbol{p})$ (one-hot distribution) and $\boldsymbol{S}(\boldsymbol{p})$ are the \textit{raw label} and predicted probabilistic distribution on $\boldsymbol{p}$, $\boldsymbol{m}$ is the weighted sum of the 
two 
distribution, and $Z$ is the scaling constant factor determined by $\pi_{rl}$, set as $Z=-(1-\pi_{rl}) \log (1-\pi_{rl})$ for convergence acceleration~\cite{englesson2021generalized}.



\textbf{3D Regularization Loss. }
In addition to the aforementioned loss on 2D renderings, 3D regularization based on our multi-scale fused features is added for further boosting the compactness of 3D Gaussian segmentation field. 
Concretely, we enforce the top K-nearest 3D Gaussians to be close in feature distances, measured by  
KL divergence, which takes the formulation as:
\begin{equation}
    \ell_{reg} = \sum_{n} \frac1{MN} \sum_{j=1}^M \sum_{k=1}^K 
    D_{\mathrm{KL}} (\boldsymbol{\mathit{f}}_{n,j} \| \boldsymbol{\mathit{f}}_{n,k})
\end{equation}
where $n$ denotes the effective feature scale ($n=3$ and $4$ 
according to our experiment),
$\boldsymbol{\mathit{f}}_{n,j}, j=1 \cdot \cdot \cdot M$ contains the $M$ sampled 3D Gaussian's fused feature from the $n$-th scale, $\boldsymbol{\mathit{f}}_{n,k}, k=1 \cdot \cdot \cdot K$ represents features of its $K$ nearest neighbors in 3D spatial space, which are activated by Softmax. 

Therefore, \textit{CoSeg} loss
in Segmentation Learning Stage is summarized as:
\begin{equation}
    \ell_{CoSeg} = \lambda_{pix} \ell_{pix} + 
    \lambda_{reg} \ell_{reg}
\end{equation}

\section{Experiments}

We evaluate our method on 
the scene
semantic segmentation task, as well as feature field reconstruction to show the effectiveness of our unprojection technique.
More results and potential applications are in \textit{Supplementary Material}.

\subsection{Experimental Setup}

\textbf{Datasets.}
Experimental results are reported on scenes from two public datasets: Replica~\cite{straub2019replica} and ScanNet~\cite{dai2017scannet}.
For both datasets, 7 complex scenes are selected for training and evaluation.
To generate 
training 
raw labels, we employ 
MaskDINO~\cite{li2023mask} 
pre-trained on ADE20K~\cite{zhou2019semantic} 
for semantic segmentation masks,
which are 
further associated by 
DEVA tracker~\cite{cheng2023tracking} according to ~\cite{ye2023gaussian}. 
Evaluation is conducted between all poses' segmentation predictions and corresponding targets in the testset. 
It should be noticed that the annotated masks are only available in evaluation as targets and are unavailable to the model during training.

\textbf{Multi-scale Features Setup.}
In reference to feature distillation works~\cite{kobayashi2022decomposing,tschernezki2022neural}, 
DINO-ViT
~\cite{caron2021emerging}
is used
as the visual 2D foundation model to extract semantic-aware features from RGBs.
For multi-scale setting, we select outputs of 4 transformer blocks.
Before unprojection to 3D Gaussians, these features are concatenated with the expanded global feature tensor then reduced to 64 dimensions by PCA and $L_2$ normalized.
For spatial features, RandLA-Net~\cite{hu2020randla}
pre-trained on the large-scale indoor dataset 
is employed, and its encoding layers are utilized as the frozen extractor during training. 

\textbf{Implementation Details.}
Our method is implemented based on official 3DGS~\cite{kerbl20233d} implementation.
We add segmentation identity, 
generated by
our Feature Unprojection \& Fusion module,
to each Gaussian.
In training, $\lambda_1=0.8$ for Gaussian Building Stage, $\lambda_{pix}=0.8$
and $\lambda_{reg}=0.2$ for Segmentation Learning Stage. 
Adam optimizer is used for updating both Gaussian parameters and the feature aggregation network,
with the former's learning rates identical to ~\cite{kerbl20233d} and 0.001 for MLP network.
All datasets are trained for 6K iterations on one NVIDIA A800 GPU.
Once trained, segmentation identities,
which serve as the output of the learned feature aggregation network, 
are stored as the 3D Gaussian attribute for rendering and inference.

\subsection{Comparisons}

\subsubsection{Scene Semantic Segmentation}

We perform a comparative analysis on the scene segmentation, which is realized by novel-view semantic segmentation.

For segmentation performance, mean intersection over union (\textbf{mIoU}) and pixel-wise \textbf{Accuracy} are employed to gauge the 
prediction quality, 
with mIoU also measuring segmentation compactness.
The reported statistics represent the average values across all test viewpoints within each dataset, with ground-truth masks utilized as targets.
Additionally,
\textbf{FPS} (Frames Per Second) (for fair comparison, we compare the time required to inference RGB \& segmentation map, as the baselines employ joint rendering) and \textbf{Learnable Parameters} during training are also used to assess the model efficiency.
Our method is compared with the NeRF-based 3D segmentation approaches, i.e., NeRF-DFF~\cite{kobayashi2022decomposing} (distilling DINO~\cite{caron2021emerging} features), Panoptic-Lifting~\cite{siddiqui2023panoptic} 
as well as Gaussian-based methods, namely Gaussian Grouping~\cite{ye2023gaussian} and Feature 3DGS~\cite{zhou2023feature} 
(using DINO as the foundation model).
All methods are trained using the same generated labels.

\begin{table*}[!h]
    \caption{
    Quantitative comparison on zero-shot novel-view semantic segmentation.
    }
    \centering
    \begin{adjustbox}{center}
    \resizebox{1.05\textwidth}{!}{
    \renewcommand\arraystretch{1} 
    \begin{tabular}{c|cccc|cccc}
    \toprule
        ~ & ~ & ~ &  \textbf{Replica} & ~ & ~ & ~&  \textbf{ScanNet}  &  ~ \\ 
         \textbf{Model} &\textbf{ mIoU(\%)} & \textbf{Acc(\%)} & \textbf{FPS} & \textbf{\makecell[c]{Learnable \\Params(MB)}} & \textbf{ mIoU(\%)} & \textbf{Acc(\%)} & \textbf{ FPS }& \textbf{ \makecell[c]{Learnable\\ Params(MB)} } \\ 
        \hline
         
         NeRF-DFF~\cite{kobayashi2022decomposing} &60.31  & 73.75 & \ \ 0.8 & 14.7  & 56.67 &67.57 &  \ \ 0.8 & 15.4  \\ 

         Panoptic-Lifting~\cite{siddiqui2023panoptic} &\cellcolor{neworange}68.87  & \cellcolor{neworange}81.83 & \ \ 3.6 & 128.5 & \cellcolor{neworange}69.33 &\cellcolor{neworange}83.25  &  \ \ 3.1 & 144.3 \\ 

         Gaussian Grouping~\cite{ye2023gaussian} & \cellcolor{newyellow}63.84  & \cellcolor{newyellow}78.03 & \cellcolor{neworange} 17.4 & 838.0 & \cellcolor{newyellow}64.99 & \cellcolor{newyellow}78.23 & \cellcolor{neworange} 17.1 & 838.6  \\ 
         
         Featrue 3DGS~\cite{zhou2023feature} & 59.62 & 70.75 & \cellcolor{newyellow} 12.9 & 977.7 & 58.48 & 68.33 & \cellcolor{newyellow} 11.7 & 1038.6  \\ 
         
         Ours & \cellcolor{newpink} \textbf{81.75}  & \cellcolor{newpink}\textbf{89.99} &\cellcolor{newpink} \textbf{18.9} & 680.3  & \cellcolor{newpink}\textbf{78.93}  & \cellcolor{newpink}\textbf{88.36} & \cellcolor{newpink} \textbf{18.0} & 684.5 \\ 

    \bottomrule
    \end{tabular}}
    \end{adjustbox}
    \label{tab:sem_ret}
\end{table*}

\begin{figure*}[!ht]
    \centering
    \begin{adjustbox}{center}
    \includegraphics[width=1.035\linewidth]{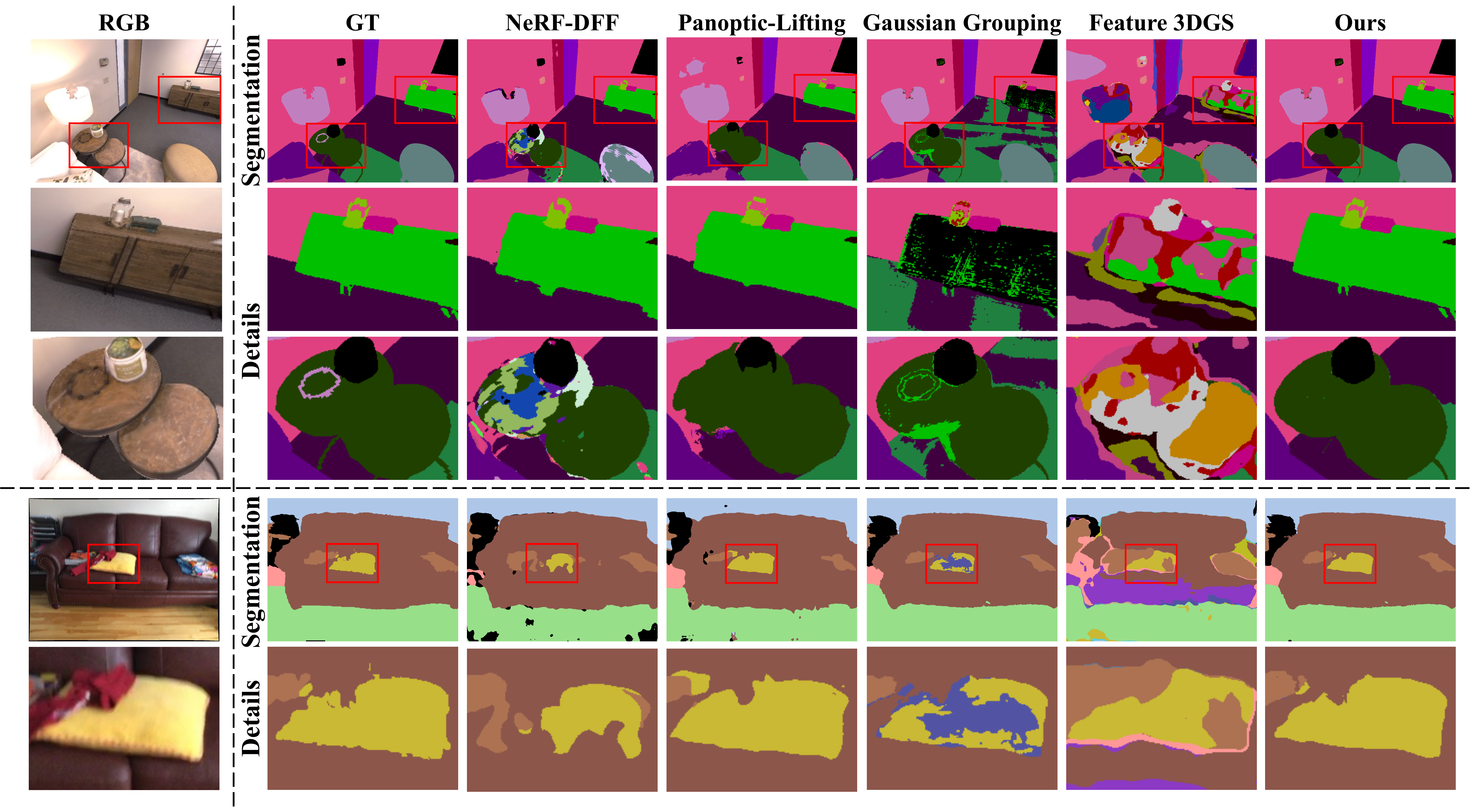}
    \end{adjustbox}
    \caption{
    Qualitative comparison on novel-view semantic segmentation.
    }
    \vspace{-13pt}
    \label{fig:sem_demo}
\end{figure*}

Tab. \ref{tab:sem_ret} demonstrates the zero-shot semantic segmentation results. 
It can be noticed our method exhibits significantly superior performance
(over +10\%/8\% mIoU on Replica/ScanNet) than the best baseline.
With less learnable parameters compared to Gaussian-based methods, the proposed method is able to mitigate the overparameterization, 
likely contributing to the enhanced performance.
Furthermore, our method delivers faster rendering speed, especially compared to NeRF-based models (roughly 20× faster than ~\cite{kobayashi2022decomposing} and 5× faster than ~\cite{siddiqui2023panoptic}).
Similar performance improvement is also showcased qualitatively in Fig. \ref{fig:sem_demo},
as our method enhances the segmentation compactness especially for large-scale objects, which probably have inconsistencies in generated raw labels. 

\subsubsection{Feature Field Reconstruction}
\label{sec:unprojection}
In order to demonstrate the effectiveness of our devised unprojection technique for introducing 2D features as the segmentation prior, 
we compare the feature field reconstructed by our approach with the distillation-based method (Feature 3DGS~\cite{zhou2023feature}).

With the feature maps rendered on novel viewpoints from the testset, comparison is conducted on the performance as well as model efficiency.
To assess the quality of the reconstructed visual semantic-aware DINO~\cite{caron2021emerging} feature field, 
Calinski–Harabasz and Davies–Bouldin Index are employed. 
Specifically, by utilizing the pixel-wise classes (provided by the test view's GT label) as the cluster identities, the aim is to ensure that rendered features could be compact inside each cluster and distinct between clusters, 
indicated by higher CH and lower DB Index.
The reconstruction time also serves as the efficiency measurement.
Additionally, comparison on LSeg~\cite{li2022language} features is also conducted to show our unprojection technique's generalizability for image-based features,  
as the visualized distribution of segmentation concepts in pixel-wise feature space are provided.

\begin{table}[!h]
    \caption{
    Comparison
    for DINO~\cite{caron2021emerging} feature field reconstruction. 
    For our method, the first time reported 
    in the parentheses 
    denotes the Gaussian building time, while the second represents the unprojection time.
    }
    \centering
    \resizebox{\linewidth}{!}{
    \renewcommand\arraystretch{2}
    \fontsize{25pt}{25pt}\selectfont
    \begin{tabular}{c| c : c c : c c}
    \toprule[3pt]
       \textbf{Method} & \hspace{1.0em}\textbf{PSNR$\uparrow$}\hspace{0.5em} & \hspace{0.8em}\textbf{CH Index$\uparrow$} & \hspace{1.5em}\textbf{DB Index$\downarrow$}\hspace{0.4em} & \textbf{Time(s)$\downarrow$} & \textbf{\makecell[c]{Learnable\\ Params(MB)}$\downarrow$} \\ 
    \Xhline{1pt}
        Distillation & 36.56 & \hspace{0.8em}695.66 & \hspace{1.2em} 4.06 & \hspace{-1em} $\approx18200$ & 1333.4 \\ 
        
        Distillation(w/ speed-up) & 36.85 & \hspace{0.8em}\textbf{699.07} & \hspace{1.2em}3.86 & \hspace{-1.2em}$\approx7400$ & 838.0 \\ 
        
        Ours & \textbf{37.41} & \hspace{0.8em}698.56 & \hspace{1.2em}\textbf{3.58} & \hspace{2.9em} $\approx \mathbf{1620}$ \Huge\textcolor[rgb]{0.7,0.8,0.9}{ (1600+20)} & \textbf{540.6} \\ 
    \bottomrule[3pt]
    \end{tabular}}
    \label{tab:unprojection}
\end{table}

        
        
\begin{wrapfigure}{r}{0.52\textwidth} 
  \centering
  \vspace{5pt}
  \includegraphics[width=\linewidth]{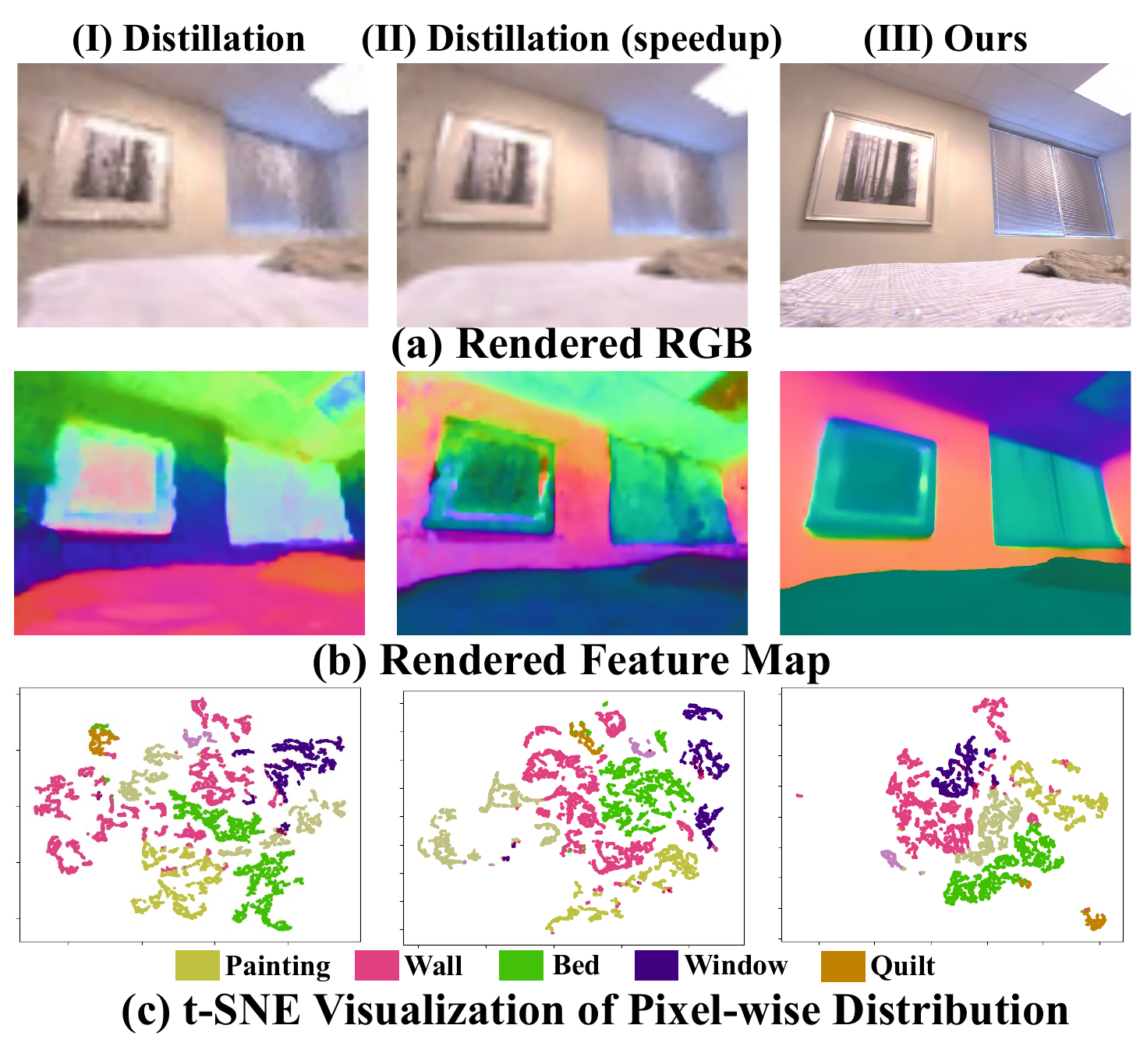} 
  \caption{    
  Visualization for LSeg~\cite{li2022language} feature field reconstruction. 
  }
  \label{fig:tsne}
\end{wrapfigure}
Tab. \ref{tab:unprojection} shows that unprojection method reconstructs the feature field with comparable or even better performance compared to the distillation-based approaches, with significantly faster speed. 
Furthermore, superior quality of rendered RGBs also illustrates the benefits of feature field reconstruction through Gaussian building + unprojection. 
It's mainly because the jointly optimized distillation will encounter the information missing caused by resolution alignment between the image and its corresponding feature map, while our decoupling approach addresses the issue, further benefiting the introduction of multi-scale feature map. 
Similarly improvement can also be found in Fig. \ref{fig:tsne}.




\subsection{Ablation Studies}
\label{ablation}
\begin{wraptable}{r}{0.525\textwidth}
    \vspace{-27pt}
    \caption{
    Ablation results on novel-view segmentation. The \textcolor[rgb]{0.7,0.8,0.9}{\raisebox{0.6ex}{\scalebox{0.7}{$\sqrt{}$}}} symbol in \textcolor[rgb]{0.7,0.8,0.9}{gray} represents the single-scale feature fusion.
    }
    \centering
    \resizebox{\linewidth}{!}{
    \renewcommand\arraystretch{1.3}
    \tiny
    \begin{tabular}{ccc|cc}
    \toprule
        \tiny \textbf{\makecell[c]{ DINO \\ Feature}} &\tiny  \textbf{\makecell[c]{Spatial\\ Feature}} &\tiny  \textbf{\makecell[c]{CoSeg \\ Loss}} &\tiny  \textbf{ mIoU(\%)} &\tiny \textbf{Acc(\%)}  \\ \hline

        × & × & × & 61.79  & 75.56  \\ 
        
        \textcolor{red}{×} & \raisebox{0.6ex}{\scalebox{0.7}{$\sqrt{}$}} & \raisebox{0.6ex}{\scalebox{0.7}{$\sqrt{}$}} & 67.53  & 79.33  \\ 
        
        \raisebox{0.6ex}{\scalebox{0.7}{$\sqrt{}$}} & \textcolor{red}{×} & \raisebox{0.6ex}{\scalebox{0.7}{$\sqrt{}$}} & 72.33  & 82.05  \\ 
        
        \textcolor[rgb]{0.7,0.8,0.9}{\raisebox{0.6ex}{\scalebox{0.7}{$\sqrt{}$}}} & \textcolor[rgb]{0.7,0.8,0.9}{\raisebox{0.6ex}{\scalebox{0.7}{$\sqrt{}$}}} & \raisebox{0.6ex}{\scalebox{0.7}{$\sqrt{}$}} & 75.69  & 84.63  \\ 
        
        \raisebox{0.6ex}{\scalebox{0.7}{$\sqrt{}$}} & \raisebox{0.6ex}{\scalebox{0.7}{$\sqrt{}$}} & \textcolor{red}{×} & 79.34 & 86.99 \\ 
        
        \raisebox{0.6ex}{\scalebox{0.7}{$\sqrt{}$}} & \raisebox{0.6ex}{\scalebox{0.7}{$\sqrt{}$}} & \raisebox{0.6ex}{\scalebox{0.7}{$\sqrt{}$}} & \textbf{81.75} & \textbf{89.99} \\ 
        \bottomrule
    \end{tabular}}
    \label{tab:ablation}
    \vspace{-8.8pt}
\end{wraptable}

\begin{figure}[!h]
    \centering    
    \begin{adjustbox}{center}
    \includegraphics[width=1.05\linewidth]{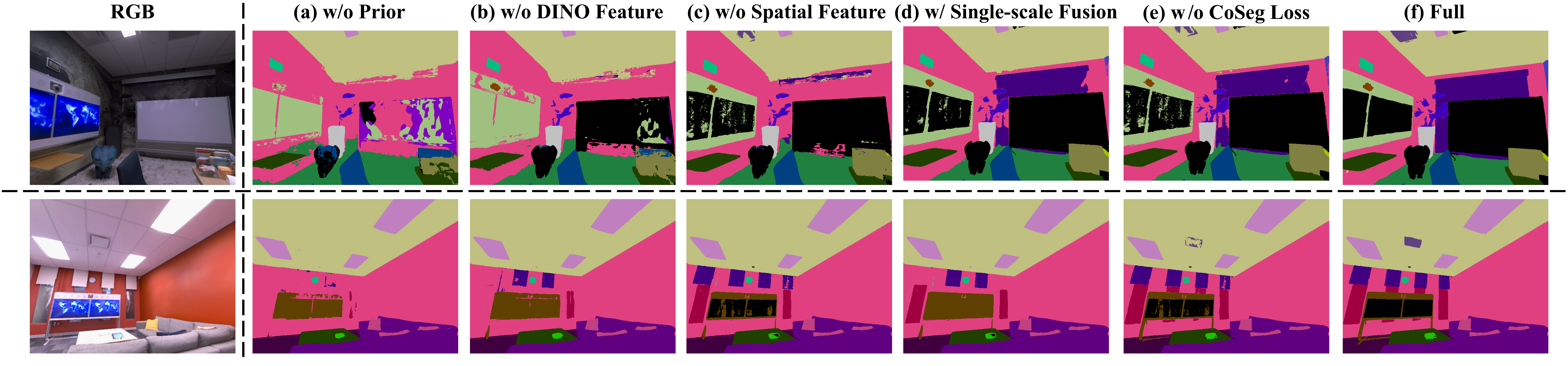}
    \end{adjustbox}
    \vspace{-20pt}
    \caption{
    Visualized ablation results. 
    }
    \label{fig:ablation}
    \vspace{-8pt}
\end{figure}

To discover each component's contribution to zero-shot segmentation improvement, a series of ablation experiments 
are conducted on Replica dataset using the same generated raw labels.
As a first baseline (Tab \ref{tab:ablation} row \textcolor{red}{1}), we disable all designed modules
and train the Gaussian segmentation identities from scratch, 
suffering from a significant performance decrease, which is also shown qualitatively in Fig \ref{fig:ablation} (a).

\textbf{DINO Feature Unprojection.} 
To verify the effect of introducing semantic-aware features, we conduct a comparison between the model without DINO unprojection 
with the full model.
Comparison between Tab. \ref{tab:ablation} row \textcolor{red}{2} and \textcolor{red}{6} reveals that 
the addition of DINO unprojection leads to remarkable improvement (mIoU +14.22\%, Acc + 10.66\%). 
Also as depicted in Fig \ref{fig:ablation} (b) and (f), the full model exhibits inferior performance, especially for segmenting large objects, like \textit{board}.

\textbf{Spatial Feature Utilization.}
The impact of Gaussian spatial features is also evaluated.
As shown in Tab. \ref{tab:ablation} row \textcolor{red}{3}, the absence of spatial information results in a performance decline (-9.42\% mIoU, -7.94\%).
Comparison between Fig. \ref{fig:ablation} (c) and (f) also demonstrates that the utilization of spatial features mainly benefits the segmentation of objects sharing similar appearance with the background, such as the \textit{projection screen} on the \textit{wall} (up) \textit{vent} in the \textit{ceiling} (below). 

\textbf{Multi-scale Feature Fusion.}
Additionally, we assess the effect of multi-scale (global-to-local) DINO and spatial feature aggregation module by 
conducting the single-scale fusion experiment.
As shown in Tab. \ref{tab:ablation} row \textcolor{red}{5} and Fig. \ref{fig:ablation} (e), the absence of multi-scale aggregation results in a performance decrease.

\textbf{CoSeg Loss.}
The effect of our proposed loss
is also evaluated, as we take comparison with the model trained using cross-entropy loss.
Tab. \ref{tab:ablation} row \textcolor{red}{5} quantitatively shows that our \textit{CoSeg Loss} will boost zero-shot scene segmentation, by 2.41\% mIoU and 3.00\% Acc, also illustrated by comparing Fig. \ref{fig:ablation} (f) with (e).


\subsection{Application}
\begin{wrapfigure}{r}{0.45\textwidth} 
  \centering
  \vspace{-30pt}
  \includegraphics[width=\linewidth]{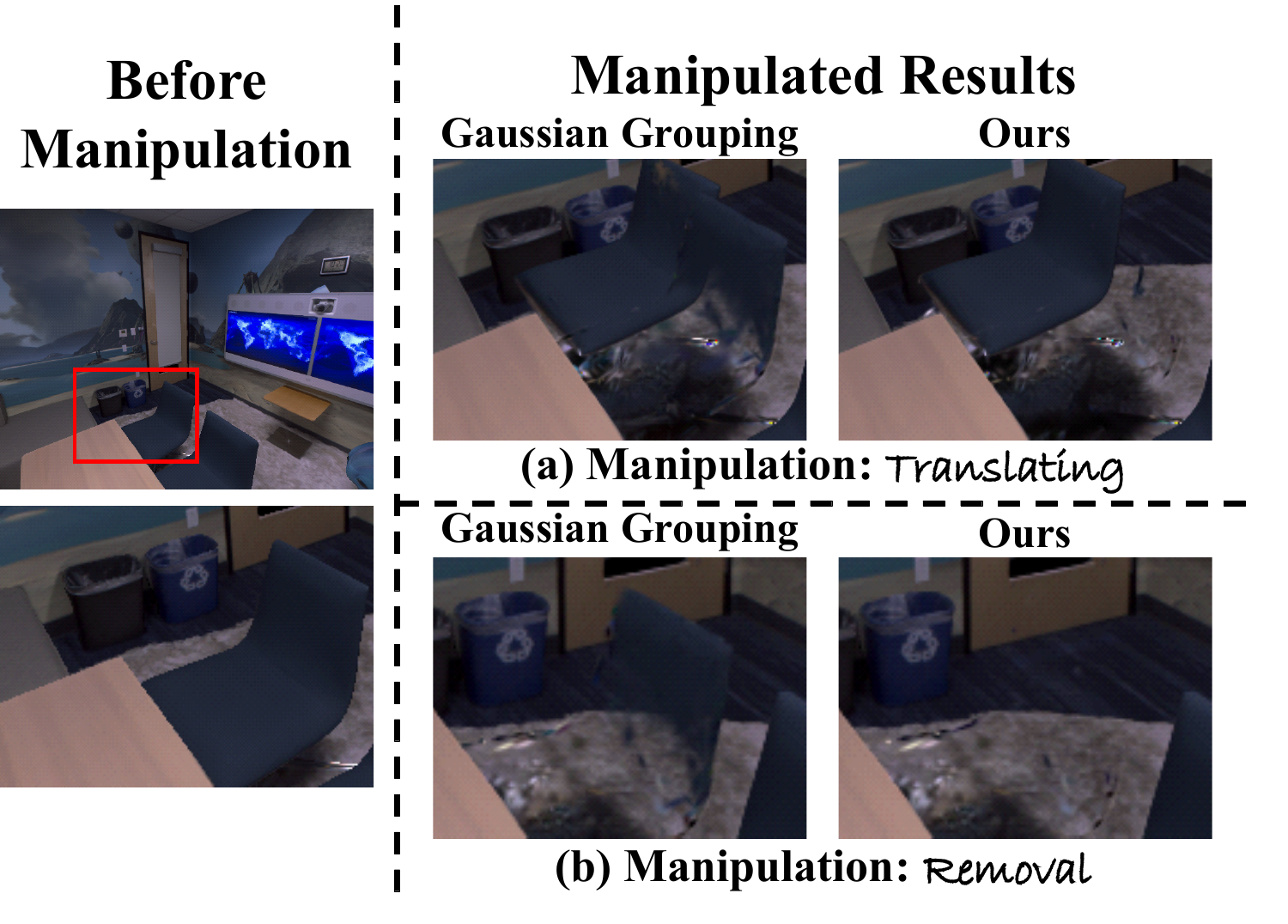} 
  \caption{    
  Scene manipulation results.
  Our model achieves better results due to the superior segmentation quality. 
  }
  \vspace{-30pt}
  \label{fig:edit}
\end{wrapfigure}
Once trained, 
in addition to achieving novel view segmentation, 
our model can also be used for 3D object-level applications 
(with the model extended for panoptic segmentation~\cite{kirillov2019panoptic}, 
results in \textit{Supplementary Material}),
such as scene manipulation. 
Fig. \ref{fig:edit} shows that
our method can achieve more realistic results for scene manipulation (no artifacts after manipulating \textit{chair}), attributed to the improved segmentation compactness.
More applications, 
such as text-prompted 3D segmentation,
are in \textit{Supplementary Material}.

\subsection{Limitations and Future Work}

Though our model requires fewer learnable parameters compared to previous Gaussian-based approaches, the GPU occupancy reaches a high level
during training to store computation graphs of the decoding MLP on Gaussian points.
This occurs despite our spatial feature extraction being designed to process only the visible Gaussians from the viewpoint instead of the whole scene, resulting in a point cloud amount that still reaches the order of magnitude of $10^5$.

Therefore in the future, we intend to make efforts to reduce GPU occupancy, for example, by reconstructing 3D Gaussians from sparser point cloud initialization. 

\section{Conclusion}

We propose CoSegGaussians, which achieves compact scene segmentation with only posed RGB images.
Our method is based on 3D Gaussian scene representation and utilizes Feature Unprojection \& Fusion to acquire the segmentation field.
For segmentation learning, semantic-aware features extracted from images across training viewpoints are gathered into the built 3D Gaussians via our unprojection techique, which are further fused with the extracted spatial information through multi-scale aggregation network to generate segmentation identities for all Gaussians. 
The network is supervised by 
the loss against 3D-inconsistent noises in raw labels.
Results and further applications
demonstrate the reliability and efficiency of our proposed method
on the zero-shot segmentation task.

%
%
%
%

\end{document}